# Temperature Distribution Prediction in Laser Powder Bed Fusion using Transferable and Scalable Graph Neural Networks

Riddhiman Raut, Amit Kumar Ball, Amrita Basak[*]

Department of Mechanical Engineering, Pennsylvania State University, University Park, PA 16802, USA

*Corresponding author
Email address: aub1526@psu.edu (Amrita Basak)

**Abstract**

This study presents novel predictive models using Graph Neural Networks (GNNs) for simulating thermal dynamics in Laser Powder Bed Fusion (L-PBF) processes. By developing and validating Single-Laser GNN (SL-GNN) and Multi-Laser GNN (ML-GNN) surrogates, this research introduces a scalable data-driven approach that learns the relevant fundamental physics from small-scale Finite Element Analysis (FEA) simulations and extends them to larger domains through knowledge transfer. Achieving a Mean Absolute Percentage Error (MAPE) of 3.77% with the baseline SL-GNN model, GNNs effectively learn from high-resolution mesh-based simulations and demonstrate strong generalization across larger geometries. The results indicate that the proposed models capture the complexity of the heat transfer process in L-PBF while significantly reducing computational costs. For instance, a thermomechanical simulation resolving scan paths in a 2 mm × 2 mm domain typically requires about 4 hours, whereas the SL-GNN model, once fully trained, can predict thermal distributions almost instantly. Additionally, calibrating models to larger domains via knowledge transfer enhances predictive performance, as evidenced by a 34.3% and 10.2% drop in MAPE for 3 mm × 3 mm and 4 mm × 4 mm domains, respectively, highlighting the scalability and efficiency of this approach. This improvement is achieved using less than 1% of the available data for retraining. Furthermore, the models show a decreasing trend in Root Mean Square Error (RMSE) when tuned to progressively larger domains, suggesting a potential for becoming geometry-agnostic. The interaction of multiple lasers complicates the heat transfer mechanism, necessitating larger model architectures and advanced feature engineering to enhance model performance. Utilizing hyperparameters obtained through Gaussian process-based Bayesian optimization (GP-BO), the best ML-GNN model demonstrates a 46.4% improvement in MAPE compared to the baseline ML-GNN model. In summary, this approach facilitates more efficient and flexible predictive modeling in L-PBF additive manufacturing.

## 1. Introduction

Metal additive manufacturing (AM) heralds a new era of design and production, characterized by its ability to forge intricate geometries that were once deemed unattainable through conventional manufacturing methods (Gao et al., 2022; Gibson et al., 2021; Hirt et al., 2017; Maleki et al., 2021; Vafadar et al., 2021; Vyatskikh et al., 2018). The essence of AM lies in its layer-by-layer construction approach. This paves the way for the creation of structures with complex internal features. Such capabilities are particularly advantageous when working with high-value materials like titanium and nickel alloy, which find extensive application in sectors demanding precision and durability, such as aerospace and medical device manufacturing. The digital foundation of metal AM facilitates unparalleled customization, making it perfectly suited for producing unique, patient-specific medical implants (Moridi, 2020; Sing et al., 2016) or small batches of aerospace components with intricate designs (Baumers et al., 2016; Blakey-Milner et al., 2021; Frazier, 2014; Thompson et al., 2016).

Laser-powder bed fusion (L-PBF) is one of the most popular AM processes, owing to its precision and material versatility (Kruth et al., 1998; Sing and Yeong, 2020; Wei and Li, 2021). The process involves spreading a thin layer of metal powder over a build platform, where a high-powered laser selectively melts the powder according to a digital design (Gu and Shen, 2009). Once a layer is fused, the build platform descends, and a new layer of powder is applied. This cycle repeats until the part is completed, layer by layer. The ability to precisely control the laser's power, speed, and focus enables L-PBF to produce parts with intricate details and tight tolerances (Yap et al., 2015). Furthermore, the rapid melting and solidification process leads to a dense microstructure with mechanical properties that can be comparable to or even superior to those of conventionally manufactured parts (Thijs et al., 2010). However, the inherent rapid thermal cycles of L-PBF can induce stresses that may distort, delaminate, or even lead to the failure of the fabricated parts. Such challenges, specifically distortion, can compromise the structural integrity and functionality of components (Zhang and Li, 2022), necessitating a thorough investigation of the fundamental physics to first understand and thereby control the manufacturing outcomes.

Distortion in L-PBF can be primarily attributed to the thermal gradient mechanism. Steep temperature gradients between the molten pool and the surrounding material cause uneven thermal expansion and contraction. When the laser heats the metal powder, the irradiated area expands but is restricted by the surrounding unheated material, developing compressive stresses. As the laser moves away, the heated area rapidly cools and solidifies, leading to shrinkage. The surrounding material, which remains at a lower temperature, resists this shrinkage, generating internal stresses and causing the top layer to curl upwards (Mercelis and Kruth, 2006; Takezawa et al., 2021). Thus, distortion in L-PBF parts is heavily influenced by temperature distribution, with the magnitude and direction of thermal gradients playing a crucial role. For example, Ali et al. (2018) demonstrated that a combination of low laser power and high exposure reduced temperature gradients, resulting in lower residual stresses. Additionally, Cheng et al. (2016) showed numerically that thermal gradients, and consequently residual stresses and distortion, are significantly affected by scanning strategies. However, these studies primarily examined small geometries, leaving it unclear whether these findings apply to larger, more complex shapes.

The introduction of multi-laser powder bed fusion (ML-PBF) systems has enabled rapid printing of large geometries owing to larger build volumes and higher build rates (Gu et al., 2021; Patwa et al., 2013); however, the presence of multiple laser beams creates irregular thermal gradients across the build area, potentially complicating stress and distortion build-up (Evans and Gockel, 2021; Masoomi et al., 2017). Mukherjee et al. (2016) demonstrated that the maximum thermal strain is directly proportional to the peak temperature achieved during printing. Zhang et al. (2020) found that using multiple lasers created a more uniform thermal environment, which reduced thermal gradients, thereby decreasing thermal stresses and part distortion. They also highlighted the importance of maintaining appropriate distances between lasers to achieve the lowest peak temperatures. Nevertheless, as with most investigations resolving scanning strategies, this study was also conducted on a small geometry, limiting the applicability of the findings to larger parts. Therefore, there is a pressing need to develop robust, generalizable frameworks that accurately predict temperature distributions as a function of scan strategies across geometries of various length scales, in order to control and mitigate thermal distortion of parts.

Currently, Finite Element Analysis (FEA) is extensively used to analyze L-PBF processes (Denlinger et al., 2014; Gouge et al., 2018). However, mesoscale simulations that compute detailed thermal distributions by resolving scan strategies are computationally expensive and time-consuming. Conversely, part-scale simulations simplify the complex thermomechanical processes using approaches such as the modified inherent strain method (Chen et al., 2019), sequential flash heating (Bayat et al., 2020), etc. While these methods keep computations tractable, they lose information about the thermal environment's evolution with laser movement. Additionally, most commercial FEA software utilizes proprietary algorithms that often lack the flexibility to incorporate custom scanning paths as inputs. This limitation hinders a comprehensive exploration of process dynamics, especially in multi-laser setups where understanding the interaction of multiple heat sources is crucial.

A thorough literature review thus reveals a substantial research gap in developing efficient and reliable predictive models for L-PBF processes. In this context, recent advancements in graph neural networks (GNNs) have shown promise in the field of mesh-based predictions. By leveraging the graph-based representation of meshes, Pfaff et al. (2020) introduced MeshGraphNets, a GNN-oriented framework tailored for simulating various problems, including cloth dynamics and structural mechanics. This approach's scalability was further investigated by Fortunato et al. (2022), who demonstrated the framework's capacity to learn dynamics on high-resolution meshes and effectively apply them to coarser, larger meshes. Such developments hint at the potential for accurately predicting complex phenomena across diverse length scales, positioning GNN-based approaches as viable candidates for developing robust predictive models for L-PBF processes.

Despite the promising advancements, scalability presents notable challenges, particularly in the context of information exchange between distant nodes, which necessitates an extensive number of message-passing steps. Addressing this issue, Gladstone et al. (2024) expanded upon the MeshGraphNet framework by introducing two innovative GNN architectures: the Edge Augmented GNN (EA-GNN) and the Multi-Graph Neural Network (M-GNN). The EA-GNN approach incorporated virtual edges to expedite information propagation, while the M-GNN adopted a hierarchical graph strategy, employing multiple graphs to focus on distinct resolutions.

These methodologies have proven successful in capturing the fundamental physics and generalizing to novel simulation scenarios, thereby demonstrating their effectiveness in simulations governed by time-independent partial differential equations (PDEs). This advancement underscores the potential of GNN-based frameworks in overcoming scalability hurdles and enhancing the precision of simulations across various scientific domains. However, addressing time-dependent PDEs, particularly transient phenomena such as heat conduction, introduces significant complexities — especially when source terms vary spatially and temporally, as observed with laser movement during layer printing in L-PBF. Furthermore, when multiple source terms are present, the model must accurately capture the physics governing the interactions among these sources, which is required for accurately predicting temperature distributions for custom scan paths in multi-laser setups.

To address the limitations identified above, this article delves into the capability of GNNs to predict the thermal fields in single and multi-laser PBF processes, given specific scanning sequences. The investigation leads to the formulation of two distinct GNN architectures — the Single-Laser GNN (SL-GNN) and the Multi-Laser GNN (ML-GNN). The ML-PBF process, involving multiple heat sources, creates thermal environments that differ significantly from those in the SL-PBF process, necessitating the creation of these two models. Accurate scaling for SL-GNN is achieved by initially transferring the knowledge of heat dissipation from smaller domains to larger geometries. The model is then fine-tuned to the new geometry through partial retraining with minimal domain-specific data. This innovative approach effectively addresses the challenge of scaling learned physical phenomena for GNNs, requiring nominal data. The introduction of multiple lasers adds significant complexity, addressed by adapting the ML-GNN's architecture alongside feature engineering and the incorporation of custom loss functions. The validity of these models is rigorously confirmed through comparisons with high-fidelity simulation data, underscoring their consistency with FEA simulations and highlighting their potential in accurately simulating heat transfer dynamics in PBF processes.

The manuscript's structure is as follows: Section 2 outlines the simulation framework and elucidates the development of GNNs. Section 3 delves into the performance evaluation of both the SL-GNN and ML-GNN models, with a focus on the efficacy of knowledge transfer and the impact of feature engineering on model accuracy. Finally, Section 4 encapsulates key insights and reflects on potential avenues for future research.

## 2. Methodology

### 2.1 Simulation methods, governing equations, domain and boundary conditions

This section discusses the simulation framework for analyzing both SL- and ML-PBF processes using Autodesk Netfabb®. The software adopts a weakly coupled, or decoupled methodology for thermomechanical modeling (Gouge et al., 2018). This methodology assumes a unidirectional influence where the thermal evolution within a part influences its mechanical properties, yet the mechanical responses exert no impact on thermal history. To encapsulate the intricacies of the L-PBF process, Netfabb integrates considerations for convective and radiative heat losses and complex thermophysical phenomena such as Marangoni convection. The model is further refined by incorporating the temperature-dependency of key thermophysical material properties, including thermal conductivity, specific heat capacity, elastic modulus, and thermal expansion coefficient, thereby enhancing the simulation's reliability. Moreover, the veracity of Netfabb's simulations has been corroborated through extensive validation by independent research groups (Gouge et al., 2019; Irwin and Gouge, 2018; Peter et al., 2020), demonstrating commendable agreement (within 5% error) with experimental observations. Consequently, the simulation outputs from Netfabb provide a high-fidelity dataset, which serves as a benchmark for assessing the predictive accuracy of the deployed machine-learning models.

### 2.1.1 Governing equations

To derive the thermal history, the transient energy balance equation is solved by transforming it into a weak formulation through the Galerkin method. Considering a part having density $\rho$, exposed to a spatially and temporally varying heat source $Q$, alongside a conductive heat flux $q$, the heat transfer equation can be written as:

$$Q(x,t) = \rho C_p \frac{dT}{dt} + \nabla \cdot q(x,t) \qquad (1)$$

Here, $T$ is the temperature of the body, $t$ is time, $q$ is the heat flux at position vector $x$ and time $t$, and $Q$ is the volumetric heat generation at $x$ and $t$. The heat flux $q$ as a function of space is given Fourier's Conduction Law:

$$q = -k\nabla T \qquad (2)$$

Here, $k$ is the thermal conductivity of the material. An initial condition, two boundary conditions, and a volumetric heat source model are required to solve the governing equation. For the initial condition, the temperature of the body is set to the ambient temperature, i.e. 25 °C. To model the heat input, convective, and radiative losses, a Neumann Boundary Condition is used, and their numerical implementation is explained in the following subsections.

### 2.1.2 Volumetric heat input model

The moving laser heat source is modeled as Goldak's 3D Gaussian ellipsoid distribution, which is given by:

$$Q = \frac{6\sqrt{3}P\eta}{abc\pi\sqrt{\pi}} \exp\left(-\frac{3x^2}{a^2} - \frac{3y^2}{b^2} - \frac{3(z+v_s t)^2}{c^2}\right) \qquad (3)$$

Here, $P$ is the laser heat source power, $\eta$ is the efficiency, $a$ and $c$ are the width and length of the ellipsoid respectively, $b$ is the depth of the ellipsoid, and $v_s$ is the laser scan velocity. The $x$ direction is normal to the motion of the heat source and the surface, $y$ is the depth of the material and the local $z$ direction is aligned along the motion of the heat source.

### 2.1.3 Thermal losses across the boundary

In L-PBF, thermal losses may occur through free convection, forced convection, and radiation. Conduction through the fixturing bodies as a source of heat loss is neglected altogether since it is common modeling practice to assume that for small geometries, most of the energy is absorbed by the substrate. Combining all the above thermal losses into a single heat transfer coefficient, the total heat flux loss can be modeled using Newton's Law of cooling:

$$q_{loss} = h(T_s - T_\infty) \qquad (4)$$

where $T_s$ is the surface temperature of the body, $T_\infty$ is the ambient temperature and $h$ is given by:

$$h = h_{free} + h_{forced} + h_{rad} \qquad (5)$$

Here, $h_{free}$, $h_{forced}$, and $h_{rad}$ are the heat transfer coefficients for free convection, forced convection, and linearized radiation respectively.

### 2.1.4 Material properties, domains, and boundary conditions

This study examines the technique of island scanning across various single-layered geometries constructed from IN625. Each domain has a thickness of 0.02 mm. Domains A (2 mm × 2 mm), B (3 mm × 3 mm), and C (4 mm × 4 mm), shown in **Figs. 1**(a)-(c) respectively, are analyzed to understand the SL-PBF process. These domains are segmented into 1 mm square islands, resulting in 4, 9, and 16 islands for domains A, B, and C, respectively. The arrangement possibilities of the four islands in domain A, calculated as 4 factorial (4!), provide 24 unique configurations, forming a significant part of the training dataset for single-laser simulations. An example arrangement for domain A, denoted by sequence [1, 2, 3, 4], is shown in **Fig. 1**(d) indicating the printing sequence of islands. Due to the vast number of possible configurations, exhaustive experimentation for domains B and C, with 362,880 and 20.99 trillion arrangements respectively, is impractical. Instead, these domains serve as cases for testing the feasibility of knowledge transfer. Selected sequences for these domains, highlighted by red dotted arrows in **Figs. 1**(b) and (c), are [7, 4, 1, 8, 5, 2, 9, 6, 3] for domain B and [13, 9, 5, 1, 14, 10, 6, 2, 15, 11, 7, 3, 16, 12, 8, 4] for domain C, demonstrating the chosen printing order.

For the ML-PBF case, a single-layered square domain of size 2 mm with a thickness of 0.02 mm is examined. This domain is segmented into three distinct areas, each assigned to one of the three lasers employed for simultaneous printing, as depicted in **Fig. 1**(e). Each laser operates within its designated area, initiating and concluding its path at any of the specified corners. The areas are filled using either a lateral or longitudinal raster scanning pattern, with **Fig. 1**(f) illustrating both pattern types. This setup forms the basis of a comprehensive full-factorial design-of-experiments (DOE), encompassing 512 simulations. This extensive dataset provides a robust foundation for in-depth analysis of the ML-PBF process. A more comprehensive understanding of the domain setup and simulation strategies can be found in (Ball and Basak, 2023a).

The temperature-dependent thermal properties of IN625, as retrieved from Netfabb's in-built library, are outlined in Table 1, while the simulation variables are listed in Table 2. The selection of these parameters has been justified in previous work (Raut et al., 2023) and has thus been excluded here for brevity.

**Table 1: Properties of IN625 as a function of temperature $T$.** Here, $K_s$ is the thermal conductivity, $\alpha$ signifies the coefficient of thermal expansion and $C_p$ denotes specific heat.

| T [°C] | k [W/mm/°C] | T [°C] | α [mm/mm/°C] | T [°C] | $C_p$ [J/g/°C] |
|---|---|---|---|---|---|
| 25 | 0.01 | 20 | 1.28e-05 | 25 | 0.405 |
| 200 | 0.0125 | 93 | 1.28e-05 | 200 | 0.46 |
| 300 | 0.014 | 204 | 1.31e-05 | 300 | 0.48 |
| 400 | 0.015 | 316 | 1.33e-05 | 400 | 0.5 |
| 500 | 0.016 | 427 | 1.37e-05 | 500 | 0.525 |
| 600 | 0.018 | 538 | 1.40e-05 | 600 | 0.55 |
| 800 | 0.022 | 649 | 1.48e-05 | 800 | 0.6 |
| 900 | 0.024 | 760 | 1.53e-05 | 900 | 0.63 |
| 1000 | 0.025 | 871 | 1.58e-05 | 1000 | 0.65 |
| 1200 | 0.0255 | 927 | 1.62e-05 | 1200 | 0.68 |

**Table 2: Variables used for FEA simulations.**

| Process Parameters | Values |
|---|---|
| Scan Speed | 1200 mm/s |
| Laser Radius | 0.05 mm |
| Power | 195 W |
| Absorptivity | 0.4 |
| Substrate temperature | 80 °C |
| Ambient temperature | 25 °C |
| Effective heat transfer coefficient | 25 W/(m² °C) |

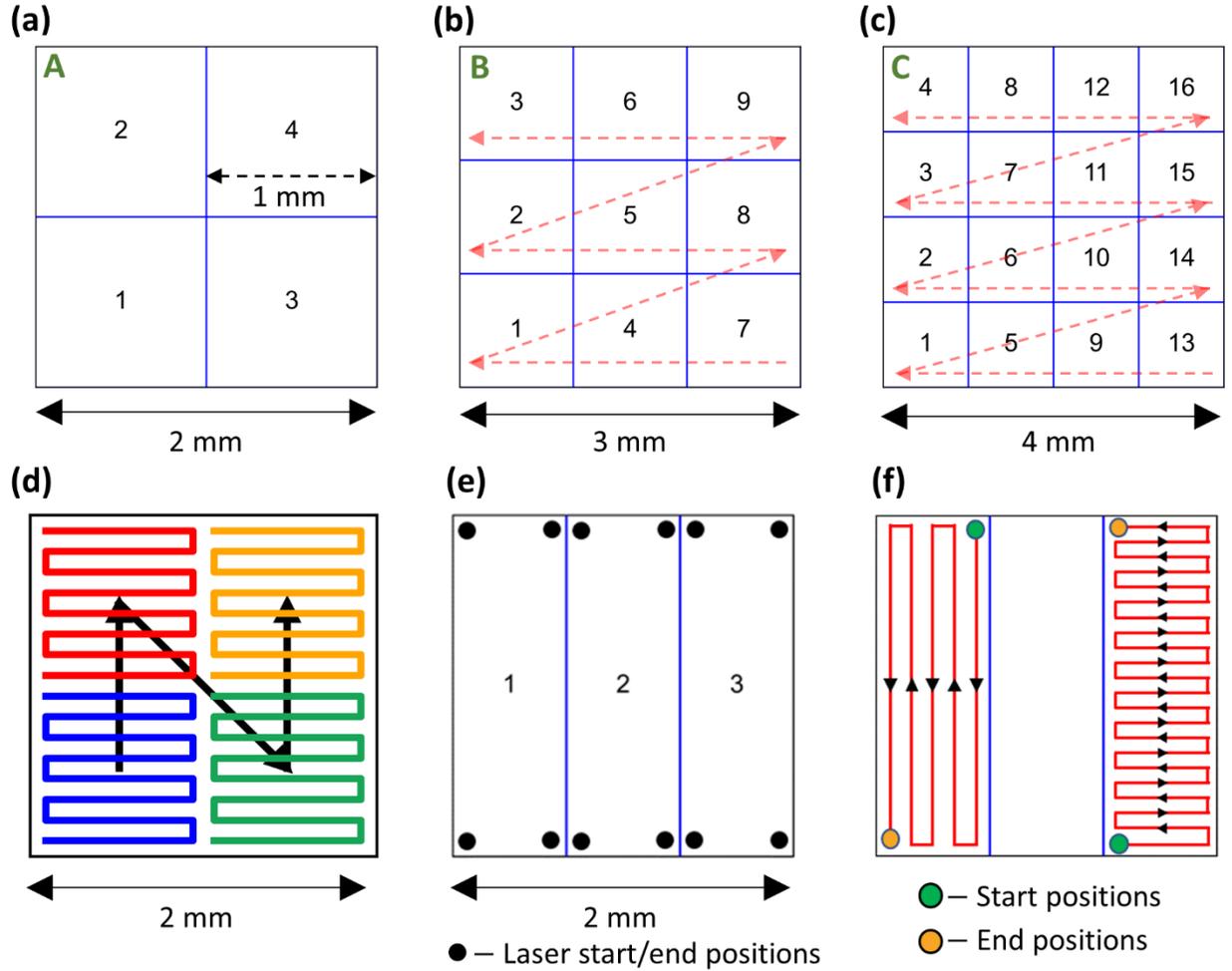

**Fig. 1.** Overview of simulation domains. The square domains of sizes (a) 2 mm, (b) 3 mm, and (c) 4 mm are studied for SL-PBF. An example printing sequence [1, 2, 3, 4] for domain A is shown in (d). The sequences selected for domains B and C are marked by the red dotted lines in (b) and (c). The domain for ML-PBF is shown in (e), with the black dots signifying possible laser start and end positions. Panel (f) demonstrates the two representative scanning strategies employed: longitudinal and lateral raster scans, which fill the designated sections.

## 2.2 Graph Neural Networks (GNNs)

### 2.2.1 Theoretical background

GNNs are a type of deep learning architecture specifically designed for graph-structured data. Meshes naturally translate to graphs, where nodes represent mesh elements and edges connect neighboring elements. This inherent graph structure allows GNNs to effectively capture the local and global relationships within the mesh (Pfaff et al., 2020).

One key advantage of GNNs for mesh predictions is their ability to learn from existing simulation data. By training a GNN on pre-computed solutions of PDEs on various meshes, the network can learn the underlying physical relationships. This knowledge can then be used to predict solutions for unseen meshes, significantly reducing computational costs compared to traditional methods (Pfaff et al., 2020).

Effective utilization of GNNs for mesh-based predictions necessitates a robust graph representation of the mesh itself. This conversion process captures the inherent connectivity within the mesh structure, transforming it into a network suitable for learning tasks, as shown in **Fig. 2**. Each element in the mesh, such as a vertex, triangle, or even a higher-order element, becomes a node in the graph. Edges are created between connected elements, typically those sharing a common boundary. This approach preserves the adjacency information crucial for capturing relationships within the mesh. Let the mesh be denoted by $M = (V, E)$, where:

- V is a set of vertices representing the mesh elements (e.g., vertices, triangles, tetrahedra).
- E is a set of edges connecting neighboring elements. An edge $(i, j) \in E$ exists if elements $v_i \in V$ and $v_j \in V$ share a common boundary.

Furthermore, relevant information $x_i \in F$ (where F is the feature space) associated with each mesh element $v_i \in V$ can be carried over as node features in the graph. These features can encompass geometric properties (e.g., coordinates), physical attributes (e.g., material properties), or even existing discrete labels. By transforming the mesh into a graph like this, a network representation encodes the structural connectivity and inherent properties of individual elements. This paves the way for GNNs to exploit these relationships for various prediction tasks within the mesh domain.

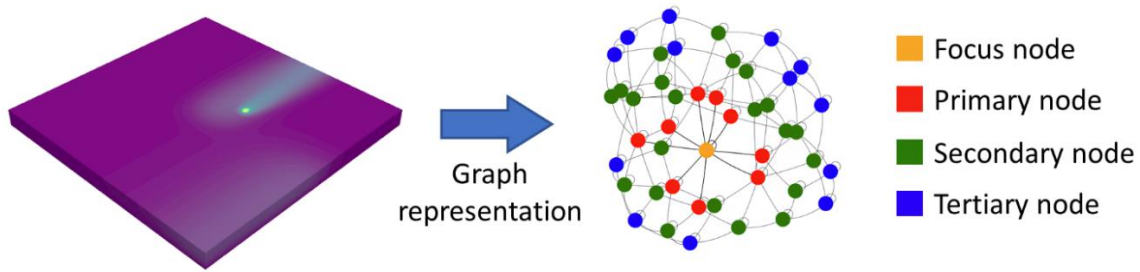

**Fig. 2.** Graph representation of a mesh. The node in focus is highlighted in orange. Adjacent nodes within one hop are marked in red and classified as primary. Nodes two and three hops away are depicted in green and blue, respectively, and are identified as secondary and tertiary nodes.

Within the realm of GNNs, distinct network architectures have been developed, each excelling in specific scenarios. One such well-established architecture is the spectral convolutional Graph Convolutional Network (GCN) introduced by Kipf and Welling (2016). GCNs leverage spectral graph theory to perform message passing, utilizing the eigenvectors of the graph Laplacian matrix to localize information propagation. This approach makes GCNs effective for tasks that rely on smooth feature transitions across the graph, such as node classification on social networks (Kipf and Welling, 2016) and semi-supervised learning on attributed graphs (Wang et al., 2020). The subsequent section offers a deeper exploration of GCNs, specifically their message-passing mechanism, and how it is leveraged for node-level predictions in mesh-based models.

**2.2.2 Graph Convolutional Networks (GCNs)**

GCNs excel at learning from graph-structured data due to their message passing mechanism, which iteratively exchanges information between neighboring nodes within the graph (Kipf and Welling, 2016; Li et al., 2024). The core idea behind GCNs is to generalize the convolution operation from grid data (like images) to graph data. In images, convolutional filters slide over local regions of the image to capture patterns. In graphs, the convolution operation is defined to aggregate information from a node's neighbors. This section explores the mathematical formulation of message passing in GCNs and its application for node-level predictions in meshes. Consider a mesh M represented as a graph $G = (V, E)$, where $V$ is the set of nodes representing mesh elements and $E$ is the set of edges connecting neighboring elements. Let $X \in R^{N \times D}$ be the node feature matrix, where $N$ is the number of nodes and $D$ is the number of features per node. The basic operation of a GCN layer can be mathematically represented as:

$$H^{(l+1)} = \sigma\left(\widehat{D}^{-\frac{1}{2}} \widehat{A} \widehat{D}^{-\frac{1}{2}} H^{(l)} W^{(l)}\right) \qquad (6)$$

In this framework, $H^{(l)}$ represents the node feature matrix at the $l$-th layer, with $H^{(0)}$ corresponding to the initial node features $X$. The weight matrix for the $l$-th layer is denoted by $W^{(l)}$. The augmented adjacency matrix, $\widehat{A} = A + I_N$ incorporates self-loops through the addition of the identity matrix $I_N$, ensuring the inclusion of each node's own features in the aggregation process. The diagonal degree matrix of $\widehat{A}$, $\widehat{D}$, is computed such that each diagonal entry, $\widehat{D}_{ii}$ equals the sum of the $i$-th row of $\widehat{A}$. The non-linear activation function, denoted as $\sigma$ (e.g., Rectified Linear Unit or ReLU), is applied to the aggregated and transformed features to introduce non-linearity into the model.

A clear view of the node-level operations in graph convolutions is presented in **Fig. 3**. The primary goal is to update the feature representation of a node $v$, depicted by the orange dot, considering the features of its neighboring nodes $N(v)$ (shown as red dots) along with its own features. Each node $v$ has an associated feature vector $x_v$. During the convolution process, these features are linearly transformed using a weight matrix $W$, which is shared across all nodes. This transformation is analogous to applying filters in traditional CNNs. The transformed features of the neighbors are then aggregated to form a single vector that captures the local topological structure. Here, the aggregation function used is the mean function, which averages the features of the neighboring nodes. This method reflects the network's ability to adapt to the graph's topology, striking a balance between simplicity, scalability, and generalizability.

The aggregated neighborhood information is then combined with the node's current feature representation, often followed by a non-linear activation function σ (in this case, ReLU). The new feature representation $x_v'$ for node $v$ can be expressed as:

$$x_v' = \sigma\big(W \cdot MEAN(x_u : u \in N(v) \cup v)\big) \qquad (7)$$

Stacking multiple graph convolution layers allows the network to capture higher-order neighborhood information. With each additional layer, a node's representation begins to encapsulate not just information from its immediate neighbors but also from nodes further away in the graph. This multilayer approach enables the model to effectively learn from graph-structured data, making graph convolutional networks (GCNs) versatile tools for tasks such as node classification, graph classification, and link prediction across diverse domains like social networks, molecular chemistry, and recommendation systems.

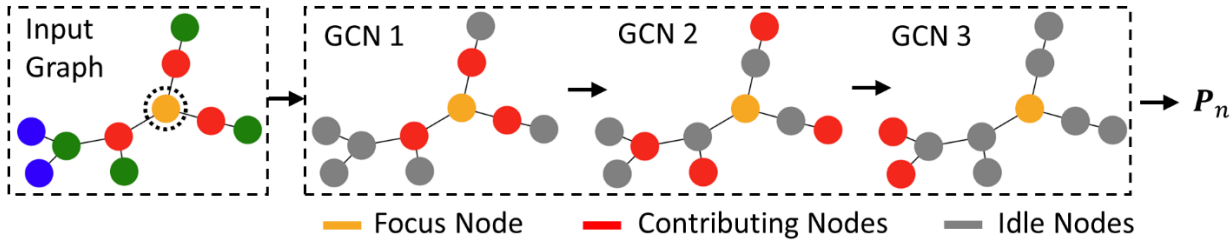

**Fig. 3.** Message passing in graph convolutions. Each successive layer of the model integrates information from increasing distances relative to the focus node to compute nodal predictions $P_n$. This process effectively captures the spatial dynamics of the nodal variables across the network.

### 2.2.3 Training methodology

#### 2.2.3.1 Single-Laser GNN (SL-GNN)

Two distinct model training approaches have been explored in this paper – complete training from scratch and employing knowledge transfer, as shown in **Fig. 4**. Initially, a GNN is trained from the ground up using simulation data from the 24 unique configurations (referred to as Cases) within domain A. The GNN architecture features four hidden layers with dimensions of 32, 64, 32, and 1, respectively, as shown in **Fig. 4**(a). To introduce non-linearity, the output from each layer is passed through a ReLU activation function. Additionally, to mitigate overfitting, a dropout layer with a rate of 0.1 is applied following each activation function.

Due to memory constraints, the entirety of this dataset is not loaded onto the GPU at a time – first, the model is trained on Case 1, then retrained using Case 2 and so on and so forth, all the way up to Case 20. Each case has 1,477 data points, which are shuffled to remove any sort of bias during the training process. These data points represent graphs, with node features including cartesian coordinates $\bar{x}$, the temperature distribution from the preceding timestep $T_{t-1}$, node type (0 for boundary and 1 for internal nodes), and a one hot-encoded laser position vector indicating the laser's focal node at the current timestep, as demonstrated in **Fig. 4**(b). The output of this network is the nodal temperature at the current timestep $T_t$ (**Fig. 4**(c)). Identifying node types is crucial because heat transfer mechanisms differ by node type—primarily convection at boundaries versus conduction within the mesh. This information is critical for the GNN to accurately learn and generalize the heat transfer dynamics across different domains. This method mirrors transient solver strategies, where solutions from previous time steps serve as initial conditions for subsequent analyses. Such training not only equips the GNN to understand the FEA process but also ensures it remains unbiased towards specific scan paths, enhancing the model's ability to predict temperature distributions for new, unseen scan paths and significantly improving its scalability and robustness.

For every case, 70% of the 1,477 data points are allocated to model training, 10% to model validation, and the final 20% for testing the model's performance. The Mean Squared Error (MSE) loss function is used to train the model and is described by the following equation:

$$\text{MSE} = \frac{1}{N}\sum_{i=1}^{N}(y_i - t_i)^2 \qquad (8)$$

Here, $N$ is the number of nodes, $y_i$ and $t_i$ represent the predicted and true temperatures for the $i^{\text{th}}$ node. This fully trained model, henceforth termed the 'FT model,' undergoes evaluation on Cases 21-24. These cases represent island sequences that the model has not previously encountered, thereby assessing the model's predictive accuracy on new data.

Employing the FT model as a foundational 'parent' model, the efficacy of transferring the learned knowledge onto larger domains is then explored, a schematic of which is shown in **Fig. 4**(d). In domain B, 20 data points out of a possible 3,210 are selected at random for this experiment. Out of these, 14 data points are allocated for training a new model—designated as TL3—where the final two layers are frozen to leverage the pre-trained features. For validation, 2 data points are utilized. A similar approach is adopted for domain C, where TL3 serves as the basis. Maintaining the last two layers frozen to preserve learned features, the model is trained using only 4 data points. An additional data point is set aside for validation, resulting in a new iteration of the model, named TL4. Thus, the models are fine-tuned to their respective domains without necessitating complete retraining or requiring extensive data.

### 2.2.3.2 Multi-Laser GNN (ML-GNN)

The ML-GNN model undergoes training with a subset of 20 cases, randomly selected from the dataset comprising 512 configurations. This methodology parallels the training strategy used for the SL-GNN model, yet it introduces several crucial adjustments. Notably, the introduction of multiple heat sources escalates process complexity by altering the domain's thermal dynamics, particularly when lasers operate close to. This complexity reduces the efficacy of a straightforward one-hot encoded laser vector, as the correlation between the number of focus nodes in the laser position vector and the resultant thermal peaks becomes non-linear, explored further in the discussions section. Adapting to these complexities necessitates modifications to the approach used for the SL-GNN model. First, the model architecture is expanded to include 2 more layers with 128 and 64 hidden dimensions respectively, as shown in **Fig. 5**. This enables the model to integrate temperature data from nodes further away from the focus node, which is vital for precisely capturing the interactions of multiple heat sources in multi-laser setups. Additionally, advanced feature engineering strategies, such as feature duplication and amplification, are employed to emphasize the significance of the laser position vector in influencing temperature distributions. Feature duplication is implemented by simply repeating the laser position vector $a$ times as a nodal feature during the graph's construction. Furthermore, feature amplification is achieved by scaling the one-hot encoded laser position vector by a factor $b$, effectively increasing the emphasis on these features and thereby enhancing their perceived importance in temperature distribution predictions.

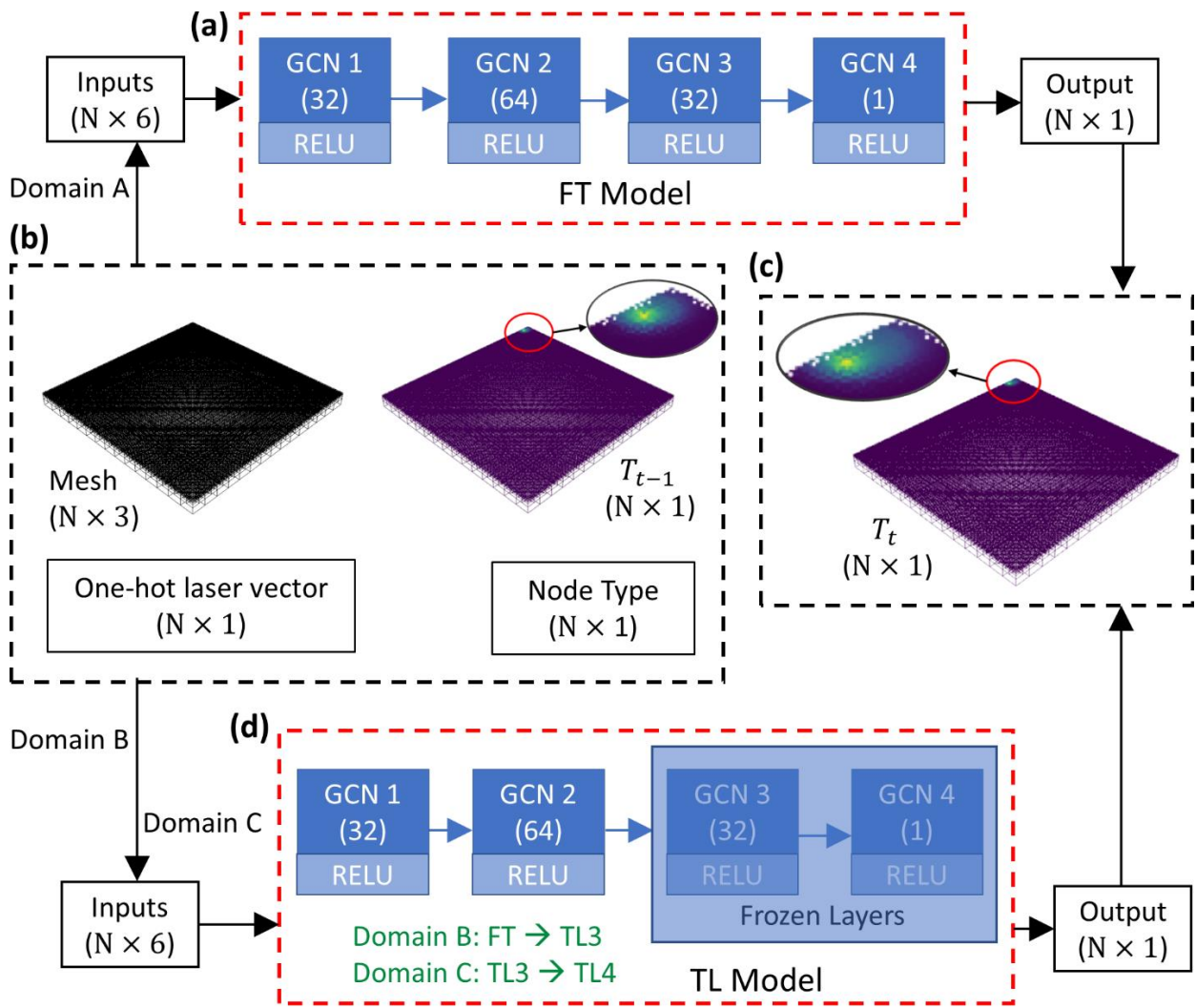

**Fig. 4.** SL-GNN training methodology. The FT Model in (a) is trained using data from domain A. Input parameters for the $N$ nodes shown in (b) include three-dimensional nodal coordinates ($N \times 3$), a one hot-encoded laser position vector ($N \times 1$), node type ($N \times 1$), and nodal temperatures from $T_{t-1}$ ($N \times 1$), thereby resulting in an $N \times 6$ input array. The output in (c) is the nodal temperature at $T_t$. The schematic for knowledge transfer is shown in (d). To create model TL3, the last two layers of the FT model are frozen while the initial two layers are retrained with a minimal amount of domain B data. Subsequently, model TL3 undergoes partial retraining with domain C data to develop model TL4.

To refine the model's predictive accuracy, particularly at temperature peaks, a dynamic loss function $L$ is adopted, as described below:

$$L = \sqrt{\frac{1}{N}\sum_{i=1}^{N} w_i \cdot (y_i - t_i)^2} \qquad (9)$$

Here, $y_i$ is the predicted temperature, $t_i$ is the target, and $w_i = c$ at the $i^{th}$ node for all temperature values greater than a threshold value of 1000 °C, and 1 otherwise. This function assigns a greater penalty to errors on peak values, magnifying their impact by a user-adjustable factor $c$. This approach ensures the ML-GNN model captures the physics governing multi-laser interactions with enhanced precision. To obtain the best possible model performance, the hyperparameter space for $a$, $b$, and $c$, are explored using Gaussian process-based Bayesian optimization (GP-BO), with a designated iteration count of 25. All simulations and data processing are conducted on a dual-processor Intel® Xeon® Gold 6230R CPU, operating at a base frequency of 2.1 GHz. Model training is performed using an NVIDIA T1000 GPU, based on the Turing architecture. A thorough evaluation of model performance, including a comparative analysis of their predictive capabilities, is elaborated in the subsequent section.

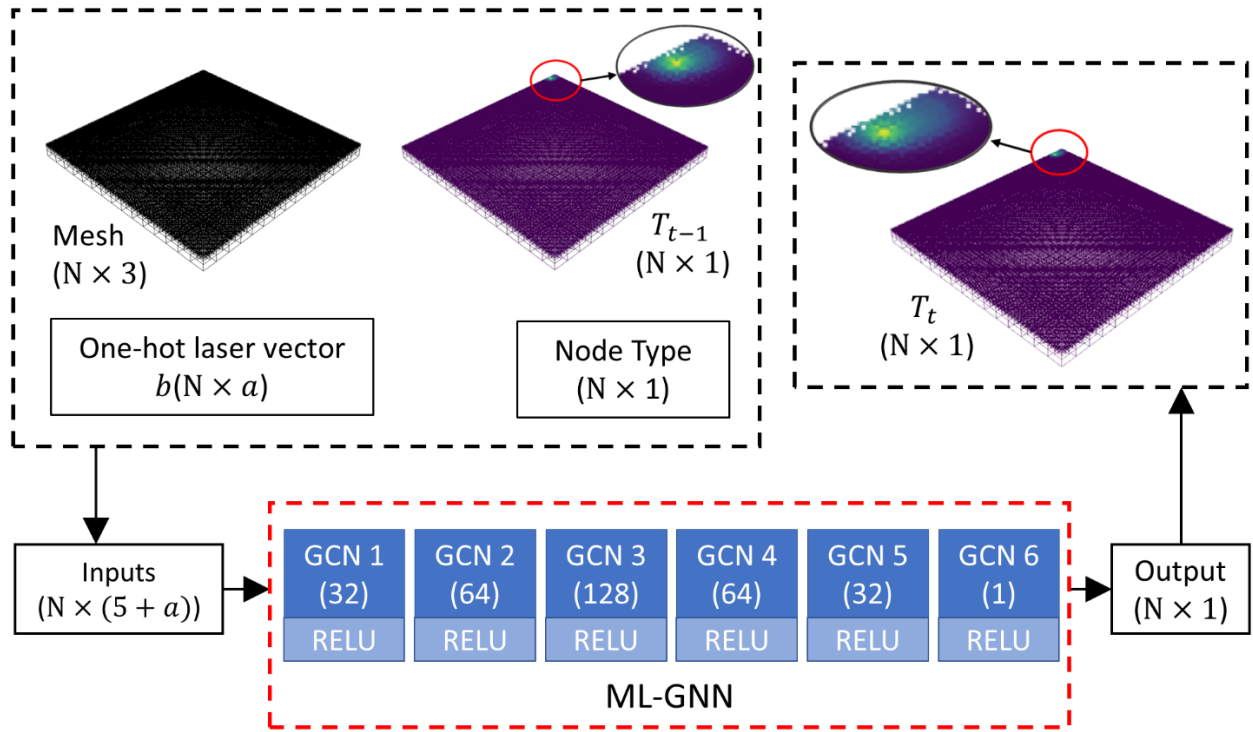

**Fig. 5.** ML-GNN training methodology. While all other input arrays remain unchanged, the one hot-encoded laser vector is duplicated $a$ times and amplified $b$ times, resulting in $5 + a$ nodal features. The final input is an $N \times (5 + a)$ matrix. The ML-GNN architecture is also expanded to include two intermediate layers with 128 and 64 hidden dimensions, respectively. This expansion enables the model to capture thermal information from more distant nodes.

## 3. Results and discussion

Part distortion in L-PBF is influenced by both the peak temperatures and the thermal gradients' magnitude and direction. Therefore, a good model must accurately capture the overall temperature distribution and peak temperatures within the domain. The following sections provide a detailed discussion of the SL-GNN and ML-GNN models' capabilities in these aspects.

### 3.1 SL-GNN

The FT model's predictive prowess in domain A is showcased in **Fig. 6**, focusing on a selected timestep $t = 450$ from Case 21. The comparison reveals a close alignment between the actual temperature distribution depicted in **Fig. 6**(a) and the model's prediction shown in **Fig. 6**(b). Notably, the model attains convergence after approximately 20,000 iterations, resulting in an MSE of around 300, as illustrated in **Fig. 6**(c). Furthermore, **Fig. 6**(d) presents a node-by-node temperature comparison at this timestep, indicating a Root Mean Squared Error (RMSE) of 16.1°C and a Mean Absolute Percentage Error (MAPE) of 3.77%. The temperature peaks are also captured with high accuracy, exhibiting a maximum Absolute Percentage Error (APE) of only 7.6%, thus exhibiting remarkable consistency with the actual data.

A detailed examination of the melt pool's trailing edge, shown in the enlarged sections of the figures, reveals a subtle distinction: the actual thermal field displays a smoother gradient compared to a slight coarsening observed in the model's prediction. However, this discrepancy is not discernible in the node-by-node temperature comparison, rendering it effectively negligible. Therefore, despite this minor variation, the model convincingly demonstrates its ability to accurately grasp and replicate the thermal dynamics, thus affirming its efficacy in mesh-level prediction tasks.

To investigate the necessity of model retraining in larger domains, the FT model is evaluated at the same timestep $t = 450$, but now in domain B, as shown in **Fig. 7**. The actual temperature distribution and its model prediction are depicted in **Figs. 7**(a) and (b), respectively. Clearly, the model accurately predicts the nodal temperatures around the laser's focal point and the melt pool, indicating that the FT model has effectively associated the laser position vector with the laser's location on any given mesh. This suggests that the model's understanding of laser positioning has generalized to arbitrary domains.

The impressive performance at the conduction region can be attributed to several key strengths inherent to GCNs. First, the graph representation of the mesh allows the FT model to efficiently handle the localized nature of heat conduction, as temperatures at each node are influenced by adjacent nodes. GCNs leverage local aggregation of information from neighboring nodes, enabling precise updates of nodal temperatures based on local thermal interactions. This local aggregation is crucial for accurately capturing the heat distribution around the laser's focal point. Moreover, through feature propagation across multiple layers, the model has learned how thermal distribution evolves over space, ensuring accurate thermal predictions for a given laser position, even when the mesh itself changes. The inherent permutation invariance of GNNs thus ensures that the model's predictions remain consistent regardless of the order of nodes in the mesh, further contributing to its robustness.

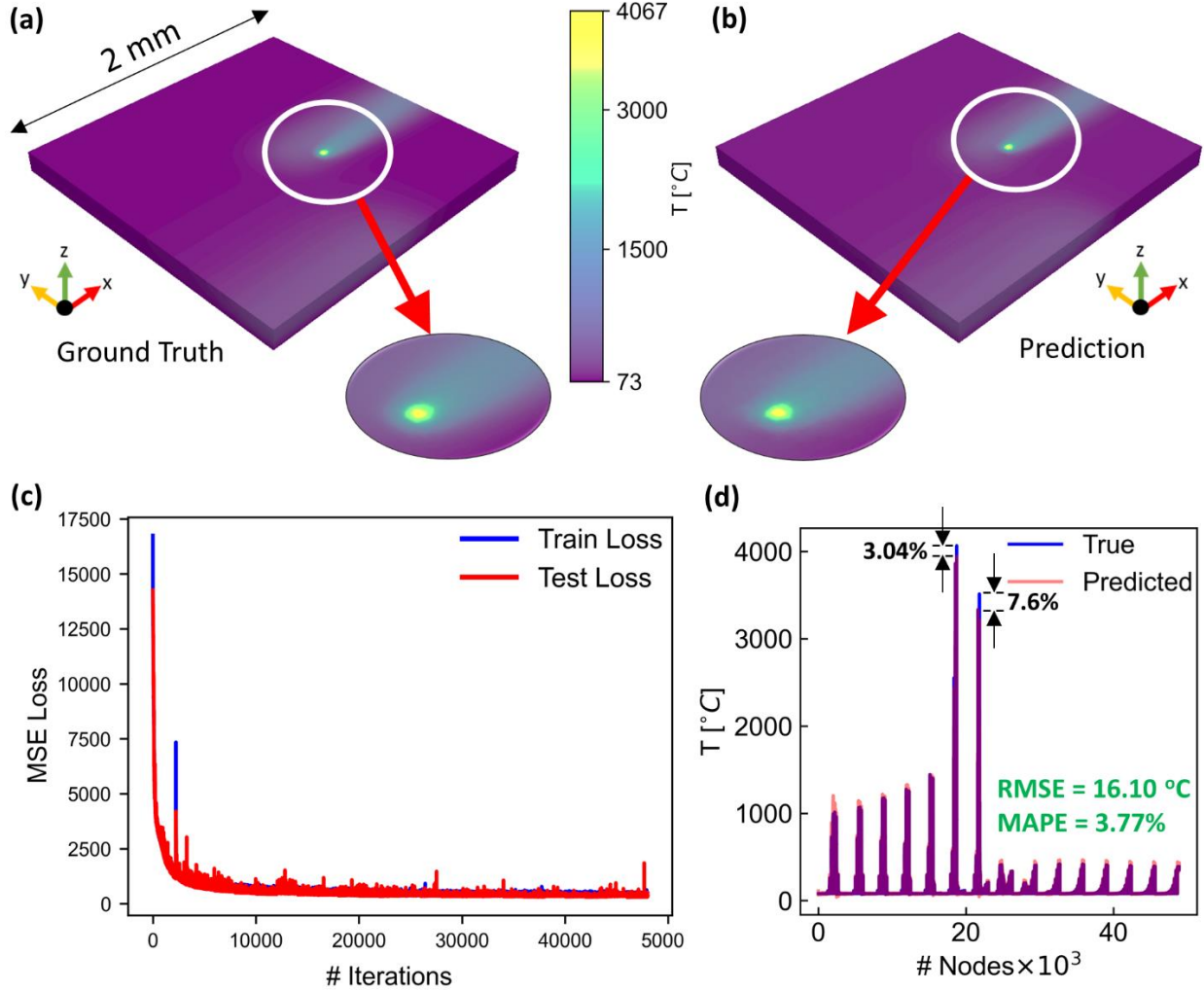

**Fig. 6.** SL-GNN performance on domain A. The ground truth temperature in (a) is in good agreement with the predicted distribution in (b). The model converges in around 20,000 iterations as depicted in (c). A node-by-node comparison of true and predicted temperatures is shown in (d).

However, a deviation arises in the thermal distribution within regions previously traversed by the laser, where heat dissipation through conduction, convection, and radiation dominates the temperature distribution. The actual temperature distribution in these zones, as shown in **Fig. 7**(c), is significantly smoother than the model's more fragmented prediction, illustrated in **Fig. 7**(d). This discrepancy is further accentuated in the node-by-node temperature comparison depicted **in Fig. 7**(e). The divergence, initially perceived as a model limitation, can be attributed to the altered thermal environment resulting from domain expansion. In domain A, with a side length of 2 mm, the laser prints closer to previously heated areas, maintaining a higher overall domain temperature and consequently slowing heat dissipation. Contrastingly, domain B is significantly larger with a side length of 3 mm. As a result, the laser often operates further from recently heated areas, facilitating quicker cooling.

This investigation yields two critical insights: firstly, the model can autonomously learn complex thermal behaviors, such as dissipation rates, without direct integration of physical phenomena via PDEs. Secondly, the model needs to be tuned to the changes in process dynamics through domain-specific training, which is further studied in the subsequent paragraphs.

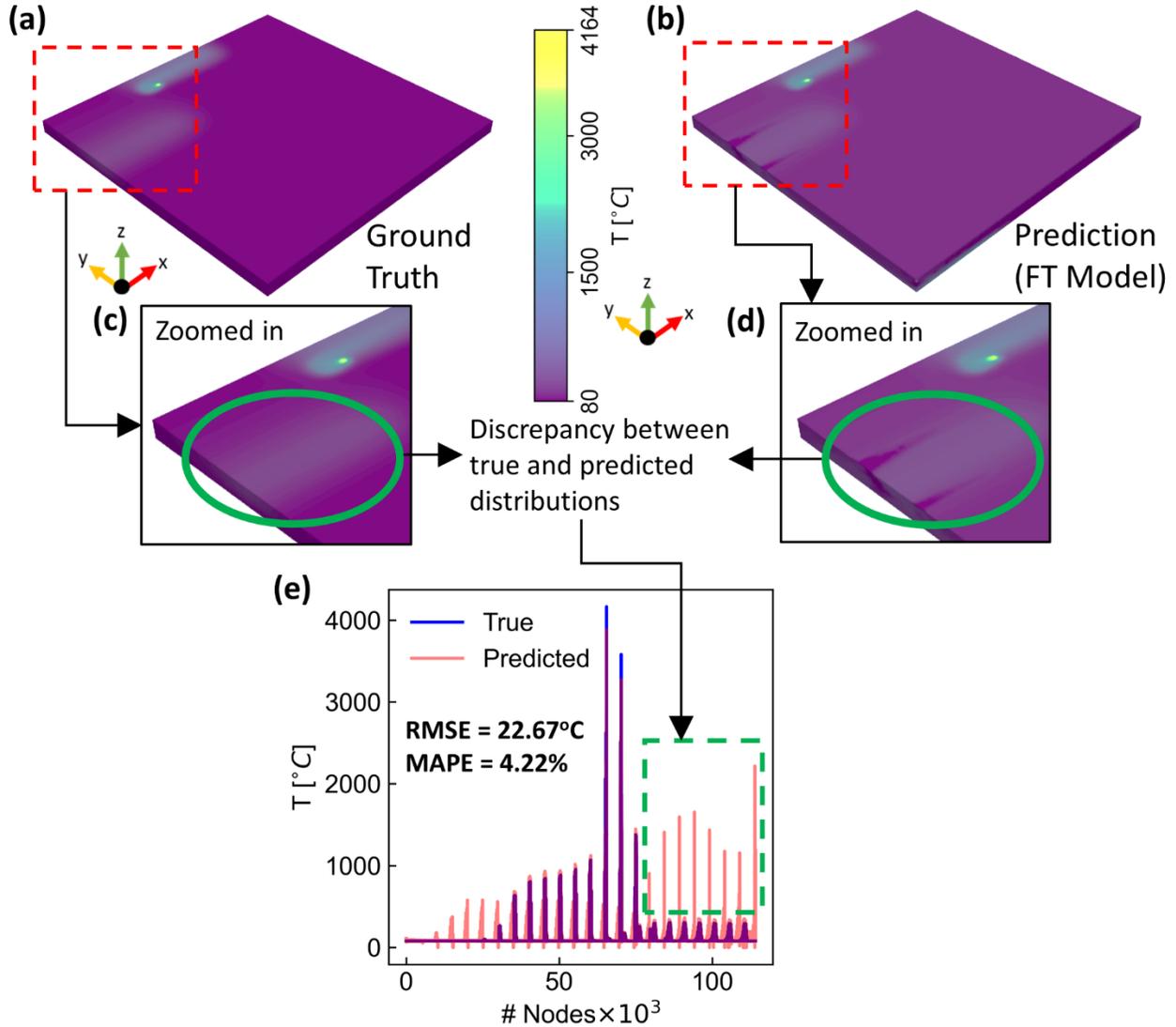

**Fig. 7.** Performance of the FT model on domain B**.** The ground truth temperature distribution in (a) agrees well with the prediction in the vicinity of the laser in (b), but exhibits differences in other regions, as shown in the zoomed plots (c)-(d). The node-by-node temperature comparison in (e) shows erroneous peaks with an RMSE of 22.67°C. Although the MAPE is quite low at 4.22%, it is skewed by the large number of nodes (over 1,20,000) compared to the 22 with spikes. These spikes suggest significant thermal peaks, undermining model fidelity.

The enhanced prediction accuracy of TL3 (obtained through partial retraining) at the 450$^{th}$ timestep is shown in **Fig. 8**. It must be noted here that this adjustment is achieved using merely 14 data points randomly chosen from a vast dataset of 3210, representing only about 0.4% of the total data. The true thermal distribution shown in **Fig. 8**(a) closely aligns with the model's prediction in **Fig. 8**(b). The enlarged views in **Figs. 8**(c) and (d) demonstrate that the areas previously impacted by the laser now display a smoother temperature distribution, consistent with high-fidelity simulation results. This enhancement is substantiated by a node-by-node temperature comparison in **Fig. 8**(e), illustrating a reduction in erroneous peaks. An improved RMSE of 14.05°C reflects a 38% enhancement compared to the method without knowledge transfer. Additionally, a reduced MAPE of 2.77% signifies a 34.3% decrease compared to the FT model's performance. These results not only demonstrate TL3's superior performance over the FT model but also validate the efficacy of the knowledge transfer approach itself. The model adeptly captures the thermal behaviors and requires only a minimal dataset to adjust to changes in geometry, underscoring the soundness of the approach.

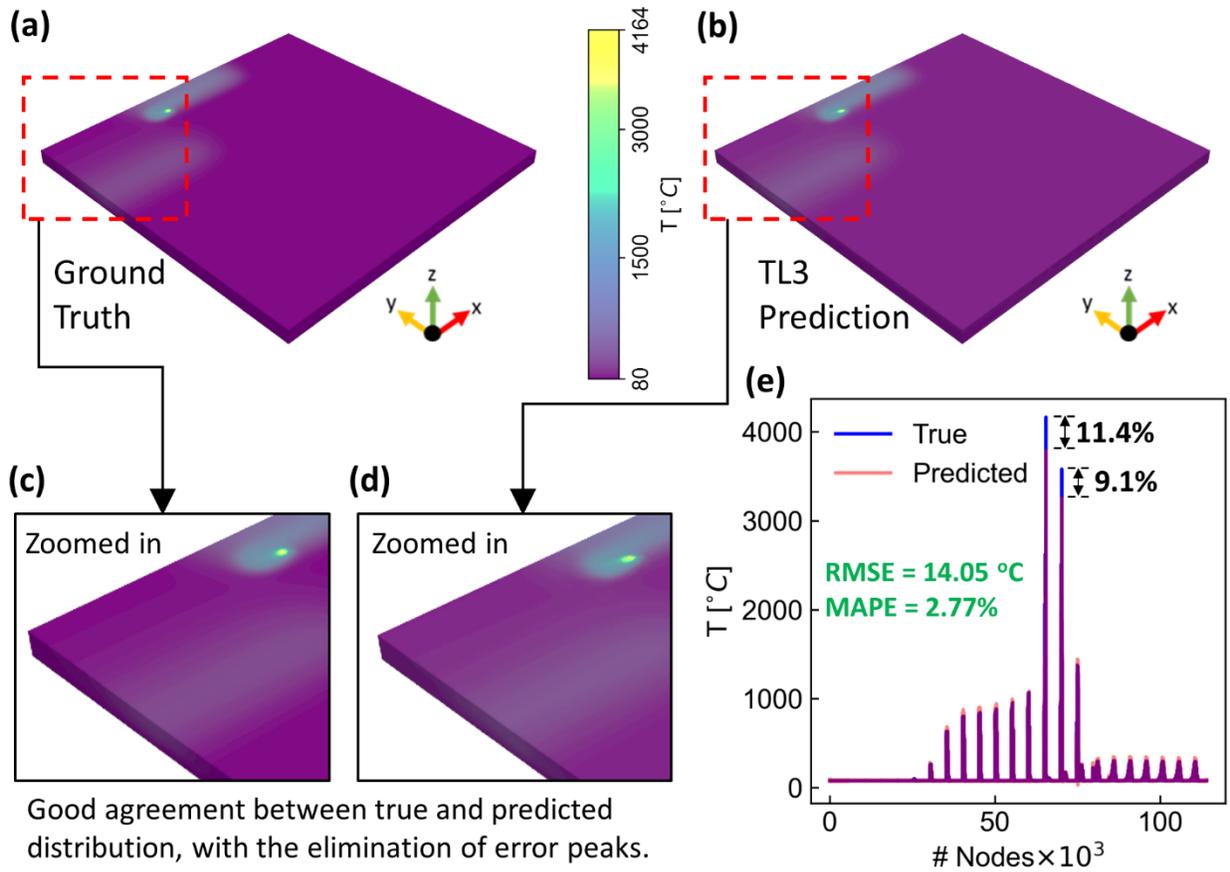

**Fig. 8.** TL3 Prediction performance on domain B. The ground truth temperature distribution in (a) is closely matched by TL3's predictions in (b). The improvements are shown using zoomed-in plots in (c) and (d). A node-by-node temperature comparison is illustrated in (e), demonstrating an improved RMSE of 14.05°C and a MAPE of 2.77%, with a maximum APE of 11.4% at the peaks.

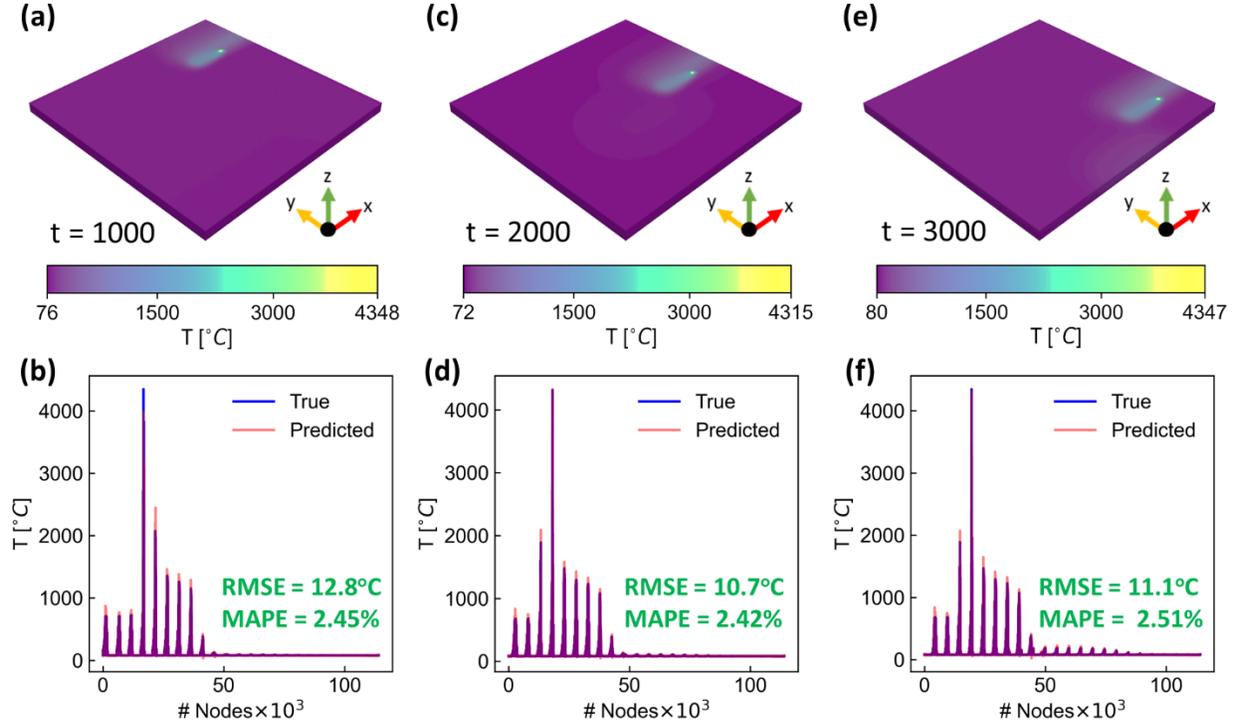

**Fig. 9.** Prediction performance of TL3 across multiple timesteps. This figure displays predicted temperature distributions and corresponding node-by-node comparisons for timesteps, $t$ of 1,000 (a)-(b), 2,000 (c)-(d), and 3,000 (e)-(f), showcasing the model's predictive accuracy at each step.

To dispel any doubts about TL3's performance being coincidental, and to affirm its genuine grasp of the underlying physics, evaluations at three subsequent timesteps – 1,000, 2,000, and 3,000 – are conducted, with the outcomes depicted in **Fig. 9**. The thermal distributions and the node-by-node temperature analyses for those timesteps are shown in **Figs. 9**(a)-(b), (c)-(d), and (e)-(f), respectively. Across these instances, the model's predictions of temperature distributions are in near-exact agreement with the actual observations, with RMSEs of 12.8°C, 10.7°C, and 11.1°C, and MAPEs of 2.45%, 2.42%, and 2.51%, respectively. Minor instances of over-prediction are observed but are deemed inconsequential, especially when considering the limited dataset employed for the model's retraining.

A noteworthy observation is the absence of error peaks in the temperature plots for these additional timesteps, a marked improvement from **Fig. 7**(e). Additionally, it is important to highlight that the original FT model is trained on a dataset spanning only 1,477 timesteps. Therefore, the transfer-learned model's exemplary performance at the 2,000[th] and 3,000[th] timesteps is not only a testament to its predictive capabilities but also to its ability to generalize the learned physical principles beyond the specific geometries or timestep ranges encountered during initial training.

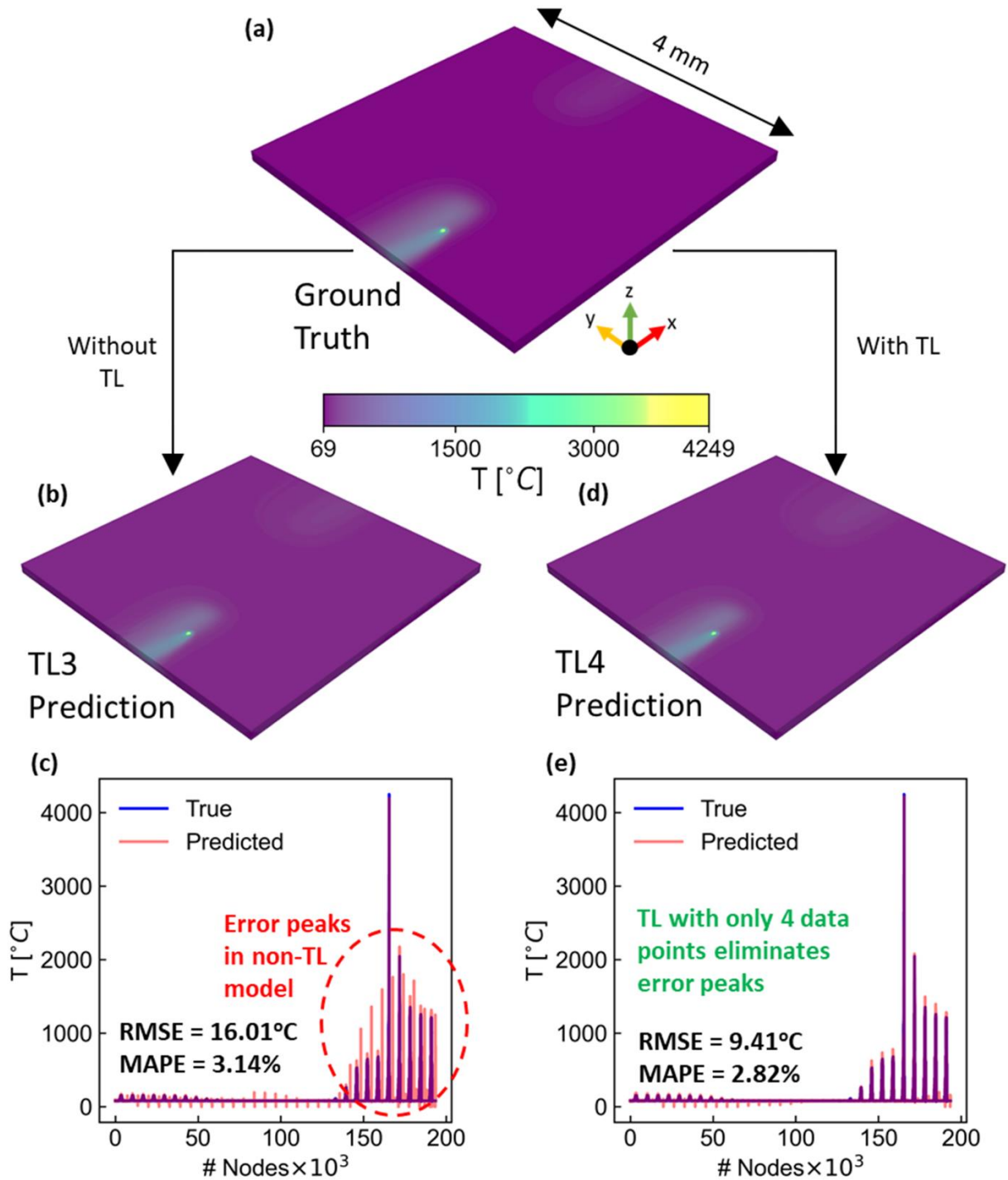

**Fig. 10.** Comparison of TL3's and TL4's predictive performance on domain C. The true temperature distribution in (a) appears to agree closely with TL3's predictions in (b), but the node-by-node comparison in (c) highlights error peaks near the laser, with an RMSE of 16.01°C and a MAPE of 3.14%. TL4's corresponding predictions in (d)-(e) demonstrate exceptional agreement, with a reduced RMSE of 9.41°C, MAPE of 2.82%, and near-exact predictions of temperature peaks.

The promising outcomes of the knowledge transfer approach raise a critical consideration: as domain sizes increase, the laser prints areas farther away from recently scanned regions, allowing these regions to cool down more. This cooling alters the laser's impact on previously scanned areas over time. By exposing the model to data from geometries of different length scales, it should ideally develop geometric independence. This evolution could significantly reduce or even negate the necessity for further retraining. To explore this possibility, the model's performance is examined at the 3,000$^{th}$ timestep in domain C, with the findings presented in **Fig. 10**. The actual temperature distribution is shown in **Fig. 10**(a), alongside the predictions made by TL3 and TL4 in **Figs. 10**(b)-(c) and **Figs. 10**(d)-(e), respectively.

Notably, TL4 is obtained from TL3 by freezing its last two layers and retraining with a mere 4 data points from domain C's extensive dataset of 5,687 points—just 0.07% of the available data. Initially, the TL3 model's predictions, while free from significant error peaks such as those observed in **Fig. 7**(e), exhibit some discrepancies primarily around the laser and melt pool areas. Although these errors are not dismissed lightly, they offer encouragement for a model that has not been directly exposed to domain C's data. The application of knowledge transfer significantly reduces these discrepancies. This is evident in TL4's predictions, which closely align with actual high-fidelity data, achieving a reduced RMSE of 9.41°C and a MAPE of 2.82%—representing decreases of 41.22% and 10.2%, respectively, compared to TL3's metrics. The elimination of error peaks with TL4 and the declining trends of RMSE and MAPE with knowledge transfer provide compelling evidence of the model's progression toward geometric agnosticism. This supports the hypothesis that exposure to larger domains and varied thermal environments enables the model to internalize complex behaviors without overfitting. Training the model on a more extensive dataset could enhance its robustness and generalizability, enabling it to be applied across arbitrary geometries without the need for further retraining.

**3.2 ML-GNN**

As touched upon in the methodology section, predicting temperature distributions become more complicated upon introduction of multiple lasers. This complexity renders the baseline architecture of SL-GNN inadequate for accurately capturing the ML-PBF process. This inadequacy is illustrated in **Fig. 11**, where panels (a) and (b) compare the high-fidelity simulations with the predictions from the baseline ML-GNN model. A noticeable discrepancy is the diminished intensity of the laser spots in the predictions compared to the actual simulations, indicating a significant underestimation of the laser-induced temperatures. This is clearly demonstrated in **Fig. 11**(c) through a detailed comparison of temperatures at individual nodes. APEs of 33.43%, 33.35%, 12.31%, and 33.07% are observed at the four distinct temperature peaks, with the maximum error highlighted in the figure. The discrepancy can be attributed to the model's inability to correlate the number of temperature peaks with the encoded laser positions, which does not ensure a direct correspondence between the presence of three lasers and the emergence of three distinct temperature peaks. Further underlining this point, as depicted in **Figs. 11**(d) and (e), is the SL-GNN scenario which results in two pronounced temperature peaks in the presence of a single laser. This makes model training more challenging and indicates the need for extensive model tuning in the form of architectural changes, feature engineering, and custom loss functions.

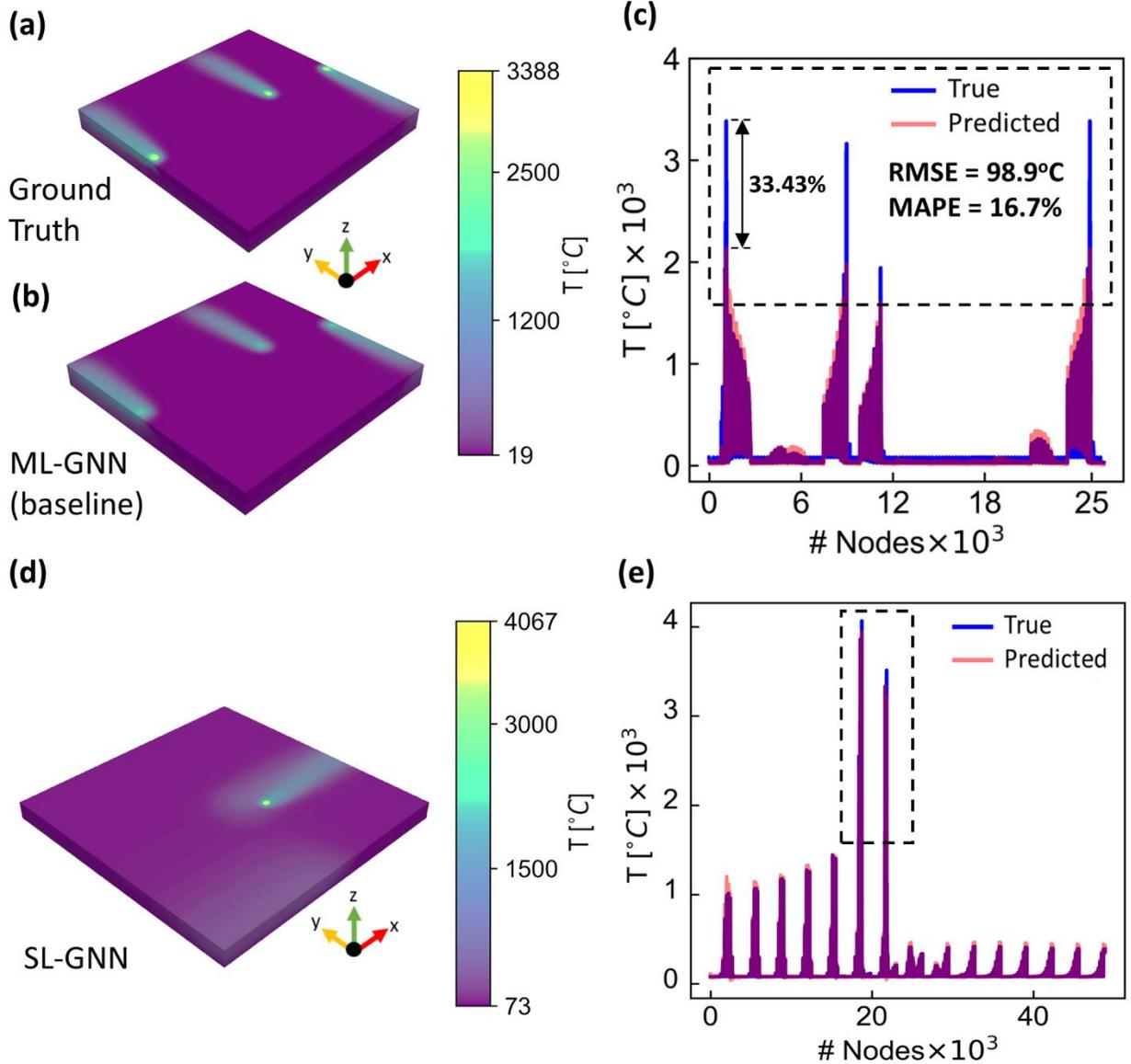

**Fig. 11.** Predictive performance of baseline ML-GNN model. The true temperature distribution and model predictions are shown in (a) and (b), respectively, while (c) presents a node-by-node comparison, indicating an RMSE of 98.9°C and a MAPE of 16.7%. The black dotted box highlights a mismatch between the three laser heads and the four temperature peaks, with a maximum APE of 33.43%. Corresponding plots for the SL-GNN case are shown in (d) and (e), respectively, underlining the superior prediction performance of the model at the peaks.

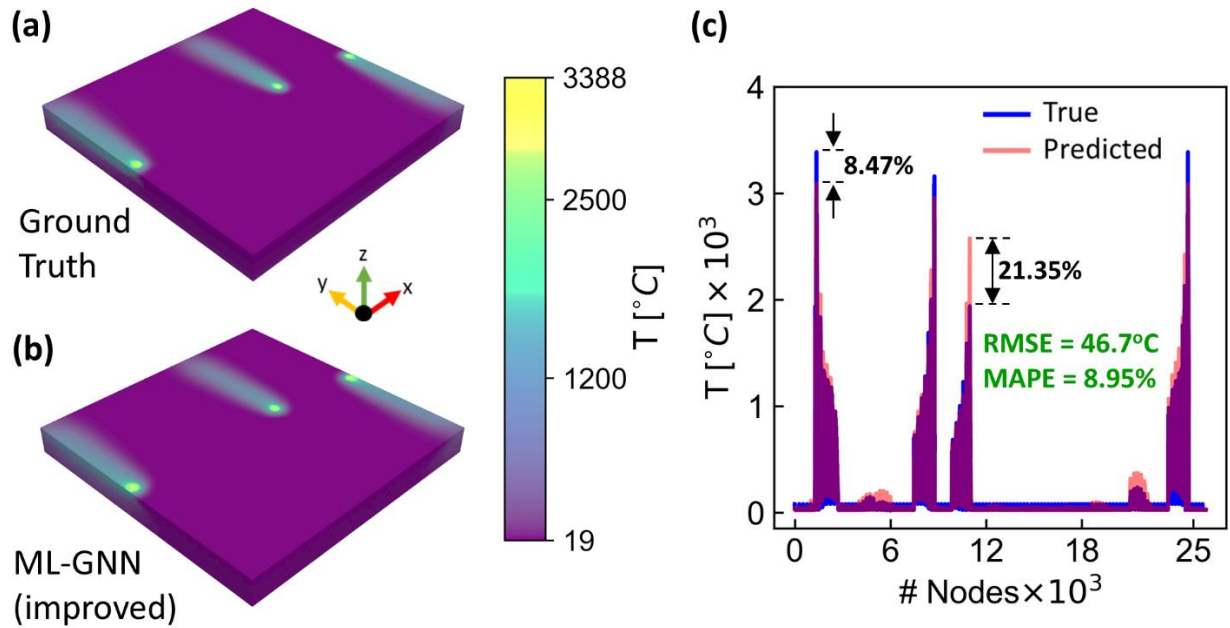

**Fig. 12.** Performance of the improved ML-GNN model. The true temperature distribution in (a) closely matches the predictions in (b). The node-by-node temperature comparison in (c) shows an improved RMSE of 46.7°C and a MAPE of 8.95%, along with more accurate predictions of peak temperatures.

The enhanced predictive performance of the improved model, characterized by hyperparameters $a = 2$, $b = 431$, and $c = 9{,}575.84$, is shown in **Fig. 12**. This model markedly surpasses its predecessor, achieving an RMSE of 46.7°C and a MAPE of 8.95%, representing reductions of 41.69% and 46.4%, respectively, compared to the baseline model's metrics. **Figs. 12**(a) and (b) juxtapose the ground truth with the model's predictions, highlighting the model's fidelity. In addition, the model significantly improves in capturing temperature peaks with APEs of 8.47%, 6.2%, and 8.46% at the laser locations (1$^{st}$, 2$^{nd}$, and 4$^{th}$ peaks), with the maximum APE of 8.47% shown in **Fig. 12**(c). Despite a 21.35% overestimation at the third peak, the model adeptly captures the overall temperature distribution. The hyperparameters of 10 of the best performing ML-GNN models obtained through GP-BO are depicted in Table 3.

**Table 3: 10 of the best-performing model hyperparameters**

| Model hyperparameters | | | RMSE [°C] |
|---|---|---|---|
| $a$ | $b$ | $c$ | |
| 2 | 226 | 8713.86 | 49.069 |
| 2 | 345 | 4828.89 | 47.427 |
| 2 | 368 | 661.46 | 47.51 |
| 2 | 404 | 6516.90 | 47.513 |
| **2** | **431** | **9575.84** | **46.7** |
| 2 | 634 | 8458.23 | 47.45 |
| 2 | 661 | 3905.38 | 49.74 |
| 2 | 831 | 4851.77 | 47.31 |

| | 2 | 872 | 8487.79 | 49.46 |

This evidence supports the use of GNNs for learning the underlying physics in complicated systems. However, such problems necessitate careful feature engineering and data preprocessing to harness the model's full capabilities. As the complexity of the task escalates, model fine-tuning becomes increasingly critical to achieve favorable outcomes.

**3.3 Thermal history predictions for a custom mesh and sequence**

In this section, the ability of the GNN to predict thermal distributions for an arbitrary user-defined scan sequence is examined. Spiral and Hilbert scan patterns (Ball and Basak, 2023b) across a 3 mm × 3 mm domain are used as inputs to the ML-GNN model. Initially, the entire domain is set to a base plate temperature of 80°C, providing the initial condition for thermal predictions at the first timestep. Subsequent temperature predictions at each timestep then serve as inputs for the next, creating a chain of predictions. The Spiral scan path for a three-laser setup is shown in **Fig. 13**(a), while **Figs. 13**(b)-(g) display the temperature distributions at timesteps 1, 100, 200, 300, 400, and 450, respectively. Similarly, the Hilbert scan path for three lasers is depicted in **Fig. 13**(h), with corresponding temperature distributions depicted in **Figs. 13**(i)-(k). The ML-GNN model adeptly correlates the one-hot encoded laser vector with the targeted laser node throughout the entire printing process, without prior exposure to any simulation data. This demonstrates that the GNN-based model effectively captures and generalizes the essential thermal physics across new domains and scanning patterns.

However, in the current methodology, the GNN utilizes only the temperature from the preceding timestep as input, making it highly sensitive to inaccuracies in predictions. The model's trajectory unrolling relies on earlier predictions to forecast future nodal variables, leading to a compounding of errors as the number of timesteps increases, which in turn diminishes the fidelity of subsequent predictions. An example of diminishing predictive performance in the spiral scan pattern is provided in **Fig. 14**: while the temperature distribution at the $5^{th}$ timestep (**Fig. 14**(a)) aligns closely with the ground truth, the predicted distribution at the $15^{th}$ timestep (**Fig. 14**(b)) struggles to accurately capture peak temperatures. The increasing trend in RMSEs for both Hilbert (dotted cyan line) and Spiral (dotted magenta line) scan patterns with an increase in timesteps, as shown in **Fig. 14**(c) further underlines the model limitations.

One potential strategy to mitigate the cascading of errors is the use of extended lookback periods. This adjustment would allow GNNs to more effectively capture the sustained heat transfer physics over longer durations. Additionally, modeling the diffusion process as time-varying edge features could enhance the model's ability to learn the heat dissipation process over time, thereby improving accuracy. Spatio-temporal GNNs (STGNNs) and Temporal Graph Networks (TGNs) offer promise, and although they have primarily been used for modeling traffic data (Tang et al., 2023) and social media feed (Rossi et al., 2020), respectively, the underlying concept remains similar. Furthermore, L-PBF involves sequentially depositing layers, effectively adding nodes to a graph over time. Consequently, it is crucial for models to learn not only the temporal variations in node features but also the changes in the graph's overall structure.

Dynamic graph models like TGNs are particularly suited for this task. They offer a robust framework for integrating structural changes over time, suggesting a promising research direction for enhancing predictive accuracy in evolving geometries.

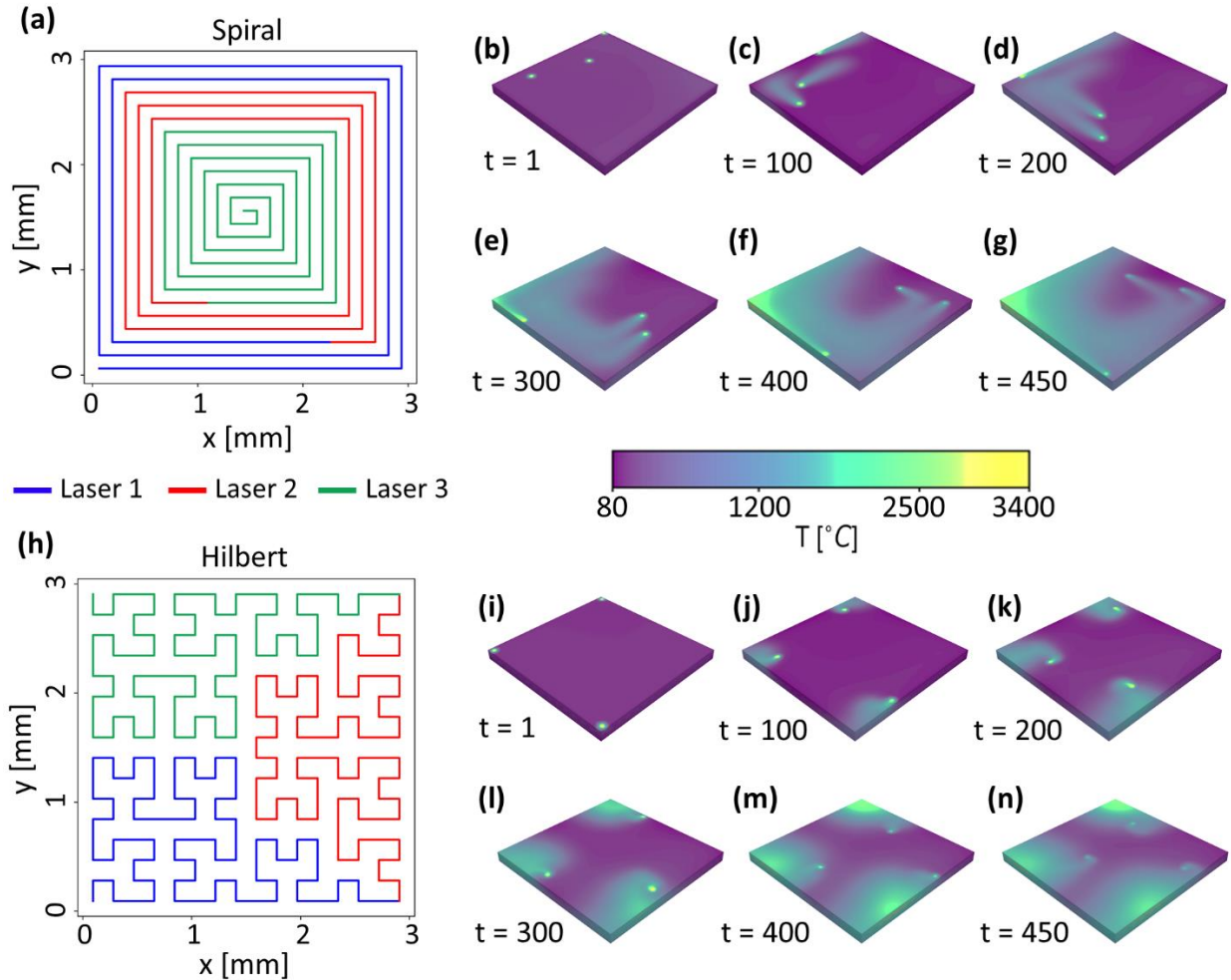

**Fig. 13.** Temperature distribution prediction for custom scan paths and meshes. Panel (a) depicts the spiral scan path for three simultaneous lasers (marked with blue, red, and green respectively) while panels (b)–(g) illustrate the predicted temperature distributions at timesteps $t$ of 1, 100, 200, 300, 400, and 450 respectively. Similarly, panel (h) demonstrates the Hilbert scan sequence, with panels (i)-(n) showing the predicted temperature distributions at the corresponding timesteps.

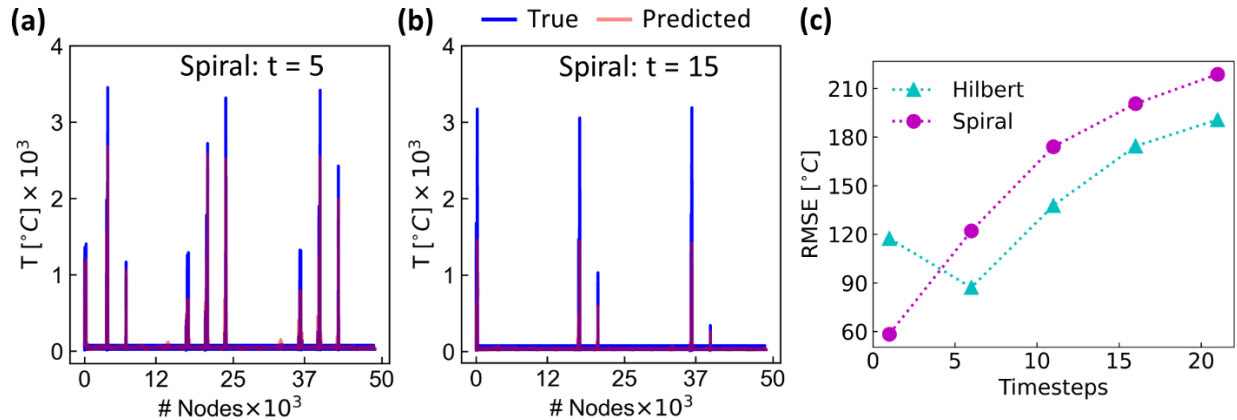

**Fig. 14.** Cascading RMSE over time. Panels (a) and (b) compare actual versus predicted temperatures node-by-node at the 5$^{th}$ and 15$^{th}$ timesteps, respectively, highlighting a decline in prediction performance. Panel (c) shows the increasing trend in RMSEs for both Hilbert and Spiral scan paths across subsequent timesteps.

## 4. Conclusion

This article explores the potential of GNNs to develop versatile models capable of predicting nodal variable distributions without incorporating any physical information. Traditional methods like FEA can facilitate highly accurate simulations only within small regions due to escalating computational demands as the domain expands. This necessitates broad assumptions to manage computational feasibility, which can lead to deviations from actual physical phenomena in larger-scale models. The novel methodology in this paper employs GNNs, trained to learn complex physical phenomena and local domain behavior, enabling the application of these models to substantially larger domains through knowledge transfer. Key findings include:

- The SL-PBF process can be precisely modeled using a straightforward 4-layer GCN, achieving a minimal MAPE of 3.77% and demonstrating strong alignment with actual simulation data.
- The model showcases promising initial predictions for larger domains, with a marked improvement in predictive accuracy due to knowledge transfer. Notably, TL3 and TL4 achieve MAPEs of 2.77% and 2.82% on domains B and C, respectively, highlighting their enhanced performance in expanded domains using smaller datasets.
- The ML-PBF process, complicated by multiple heat sources, necessitates an adapted model architecture and feature engineering to capture these complexities. The resulting ML-GNN model significantly outperforms the baseline, achieving 41.69% and 46.4% reductions in RMSE and MAPE, respectively. This improvement underscores the effectiveness of GNN-based FEA surrogates in handling complex scenarios, contingent upon feature preprocessing and hyperparameter tuning.
- Predicting entire thermal histories by using only the previous timestep's solution as input makes the model sensitive to erroneous predictions, underlining the need for incorporating extended lookbacks to more effectively capture temporal trends.

This foundational study opens various avenues for future research, such as integrating additional process parameters to enhance model generalizability across different machines and materials. The development of a comprehensive high-fidelity data library, similar to ImageNet (Deng et al., 2009) in the image processing domain, could further refine model robustness, facilitating rapid predictive capabilities. The scalability offered by this new approach will also allow FEA to delve into more intensive, physics-driven models for small-scale phenomena. The synergy between first-principles calculation (e.g., FEA) and artificial intelligence heralds a promising frontier in computational modeling, offering exciting prospects for advancing this field.

**Author contribution**

Conceptualization, A.B. and R.R.; Methodology, R.R. and A.K.B; Software, R.R. and A.K.B; Validation, R.R.; Formal analysis, R.R.; Investigation, R.R.; Resources, A.B.; Data curation, R.R. and A.K.B.; Writing – original draft, R.R.; Writing – review & editing, A.B. and A.K.B; Visualization, R.R.; Supervision, A.B.; Project administration, A.B.; Funding acquisition, A.B. All authors have read and agreed to the published version of the manuscript.

**Data availability statement**

The data presented in this study are available from the corresponding author on reasonable request.

**Funding**

The work reported in this paper is funded by the Defense Advanced Research Projects Agency [grant number D22AP00147]. Any opinions, findings, and conclusions in this paper are those of the authors and do not necessarily reflect the views of the supporting institution.

**Acknowledgments**

The authors would like to thank Jeff Irwin, Research Engineer at Autodesk, for his assistance with Netfabb, and Chris Hirsh, IT Consultant in the Department of Mechanical Engineering at Penn State, for his support with software installation and remote operation.

**Conflicts of Interest**

The authors declare no conflict of interest. The funders had no role in the study's design, the collection, analyses, or interpretation of data, the writing of the manuscript, or the decision to publish the results.

Challenges. Appl. Sci. 11, 1213. https://doi.org/10.3390/app11031213

Vyatskikh, A., Delalande, S., Kudo, A., Zhang, X., Portela, C.M., Greer, J.R., 2018. Additive manufacturing of 3D nano-architected metals. Nat. Commun. 9, 593. https://doi.org/10.1038/s41467-018-03071-9

Wang, Y., Tang, S., Lei, Y., Song, W., Wang, S., Zhang, M., 2020. DisenHAN: Disentangled Heterogeneous Graph Attention Network for Recommendation, in: Proceedings of the 29th ACM International Conference on Information & Knowledge Management. ACM, New York, NY, USA, pp. 1605–1614. https://doi.org/10.1145/3340531.3411996

Wei, C., Li, L., 2021. Recent progress and scientific challenges in multi-material additive manufacturing via laser-based powder bed fusion. Virtual Phys. Prototyp. 16, 347–371. https://doi.org/10.1080/17452759.2021.1928520

Yap, C.Y., Chua, C.K., Dong, Z.L., Liu, Z.H., Zhang, D.Q., Loh, L.E., Sing, S.L., 2015. Review of selective laser melting: Materials and applications. Appl. Phys. Rev. 2. https://doi.org/10.1063/1.4935926

Zhang, J., Li, B., 2022. The Influence of Laser Powder Bed Fusion (L-PBF) Process Parameters on 3D-Printed Quality and Stress–Strain Behavior of High-Entropy Alloy (HEA) Rod-Lattices. Metals (Basel). 12, 2109. https://doi.org/10.3390/met12122109

Zhang, W., Tong, M., Harrison, N.M., 2020. Scanning strategies effect on temperature, residual stress and deformation by multi-laser beam powder bed fusion manufacturing. Addit. Manuf. 36, 101507. https://doi.org/10.1016/j.addma.2020.101507